\documentclass[a4paper, conference]{IEEEtran}
\IEEEoverridecommandlockouts

\usepackage{cite}
\usepackage{amsmath,amssymb,amsfonts}
\usepackage{algorithmic}
\usepackage{graphicx}
\usepackage{textcomp}
\usepackage{xcolor}

\usepackage{caption}
\usepackage{subcaption}
\usepackage{booktabs}
\usepackage{balance}

\def\BibTeX{{\rm B\kern-.05em{\sc i\kern-.025em b}\kern-.08em
    T\kern-.1667em\lower.7ex\hbox{E}\kern-.125emX}}

\newcommand{\etal}{{\textit{et al}}.\@ }

\begin{document}

\title{Anomaly Detection using Deep Reconstruction and Forecasting for Autonomous Systems}

\author{\IEEEauthorblockN{Nadarasar Bahavan\IEEEauthorrefmark{1}, Navaratnarajah Suman\IEEEauthorrefmark{1}, Sulhi Cader\IEEEauthorrefmark{2}, Ruwinda Ranganayake\IEEEauthorrefmark{1}, Damitha Seneviratne\IEEEauthorrefmark{2},\\ Vinu Maddumage\IEEEauthorrefmark{1}, Gershom Seneviratne\IEEEauthorrefmark{1},Yasintha Supun\IEEEauthorrefmark{1}, Isuru Wijesiri\IEEEauthorrefmark{2}, Suchitha Dehigaspitiya\IEEEauthorrefmark{2},\\ Dumindu Tissera\IEEEauthorrefmark{1} and  Chamira Edussooriya\IEEEauthorrefmark{1}}
\IEEEauthorblockA{\IEEEauthorrefmark{1}Department of Electronic and Telecommunication Engineering, University of Moratuwa, Sri Lanka \\
\IEEEauthorrefmark{2}Department of Computer Science and  Engineering, University of Moratuwa, Sri Lanka }}

\maketitle

\begin{abstract}
We propose self-supervised deep algorithms to detect anomalies in heterogeneous autonomous systems using frontal camera video and IMU readings. Given that the video and IMU data are not synchronized, each of them are analyzed separately. The vision-based system, which utilizes a conditional GAN, analyzes immediate-past three frames and attempts to predict the next frame. The frame is classified as either an anomalous case or a normal case based on the degree of difference estimated using the prediction error and a threshold. The IMU-based system utilizes two approaches to classify the timestamps; the first being an LSTM autoencoder which reconstructs three consecutive IMU vectors and the second being an LSTM forecaster which is utilized to predict the next vector using the previous three IMU vectors. Based on the reconstruction error, the prediction error, and a threshold, the timestamp is classified as either an anomalous case or a normal case. The composition of algorithms won runners up at the IEEE Signal Processing Cup anomaly detection challenge 2020. In the competition dataset of camera frames consisting of both normal and anomalous cases, we achieve a test accuracy of 94\% and an F1-score of 0.95. Furthermore, we achieve an accuracy of 100\% on a test set containing normal IMU data, and an F1-score of 0.98 on the test set of abnormal IMU data.
\end{abstract}


\section{Introduction}
\label{se:intro}

Anomalies are unexpected or abnormal behaviour of systems, which are synonymously referred to as outliers, exceptions, peculiarities, discordant observations, contaminants, or aberrations in different contexts~\cite{chandola2009anomaly}. Anomaly detection is a crucial task in many applications, such as, fraud detection for credit card transactions, intrusion detection for cyber-security systems, illegal activity detection for surveillance systems, and fault detection in safety-critical systems. We consider the real-time detection of abnormalities of ground and aerial heterogeneous autonomous systems using their embedded sensor data. For this purpose, information given by the camera (i.e., video) and the inertial measurement unit (IMU) sensor can be exploited to detect anomalies. By employing both video and IMU data, the robustness of the anomaly detection can be improved since the information of some anomalies may be available either in video data or in IMU data, for example, sudden obstructions to a regularly moving system can be mostly detected using video data only.

\begin{figure}[t]
\begin{subfigure}[b]{\columnwidth}
\includegraphics[width=0.49\linewidth]{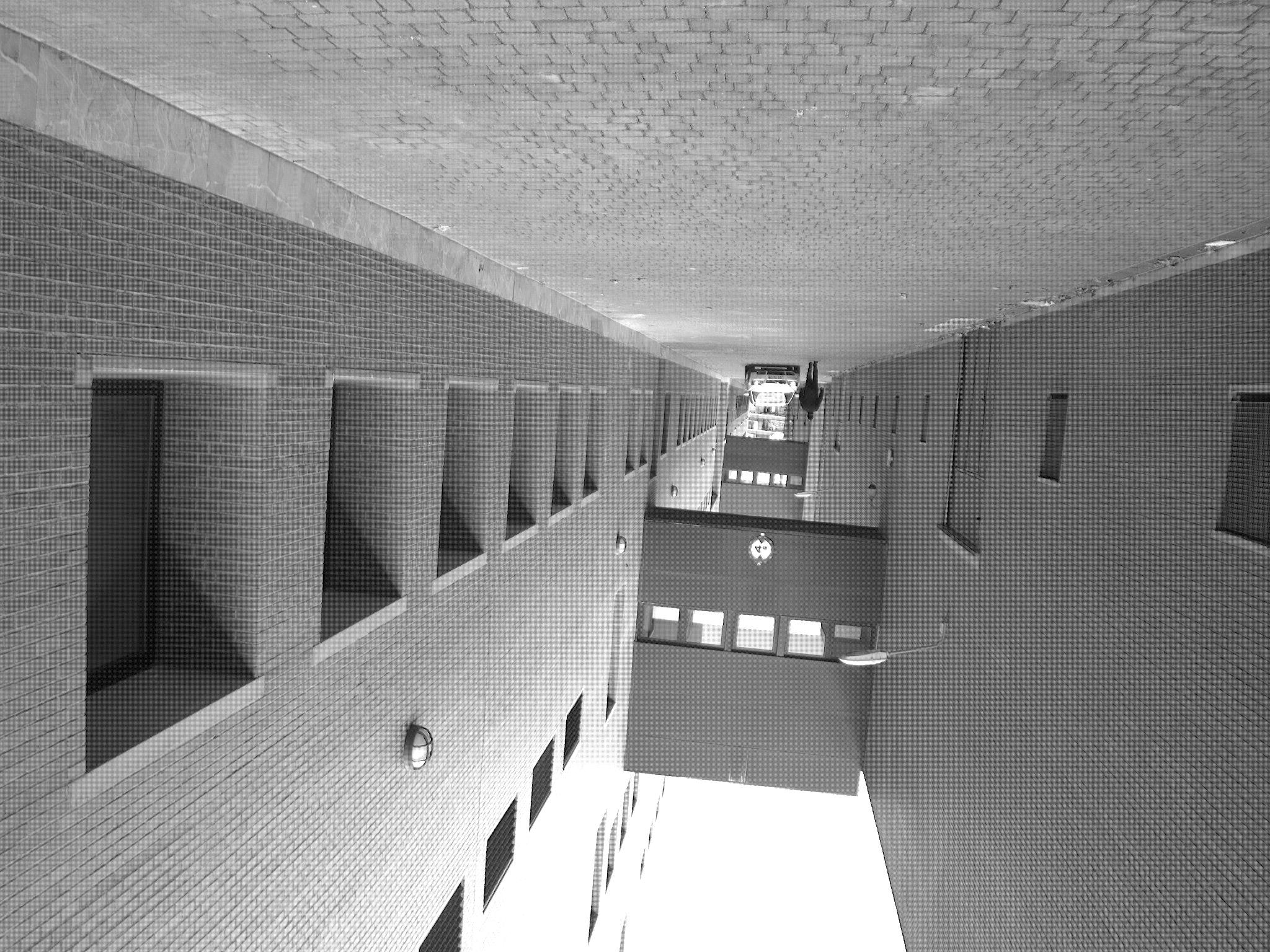}
\includegraphics[width=0.49\linewidth, height=1.4in]{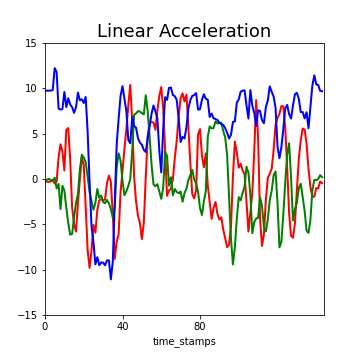}
\caption{Instant Anomalies}
\label{fig:types1}
\end{subfigure}
\begin{subfigure}[b]{\columnwidth}
\includegraphics[width=0.24\linewidth]{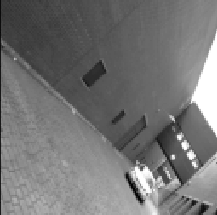}
\includegraphics[width=0.24\linewidth]{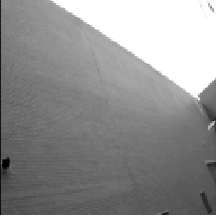}
\includegraphics[width=0.24\linewidth]{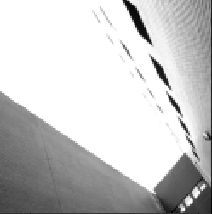}
\includegraphics[width=0.24\linewidth]{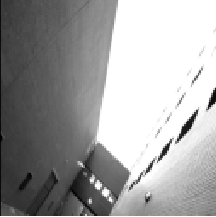}
\caption{Transitional Anomalies}
\label{fig:types2}
\end{subfigure}
\vspace{-0.05in}
\caption{(a) Instant anomalies: Anomalous instances such as spatial anomalies (left) and anomalies in linear acceleration values (right).
(b) Transitional Anomaly: Anomalies in the transition between subsequent instances such as the sequence of images shown here.}
\label{fig:types}
\vspace{-0.05in}
\end{figure}

We divide the broad range of anomalies in autonomous agents into two basic categories based on the circumstance on occurrence. They are:
\vspace{-0.04in}
\begin{enumerate}
    \item Internal anomalies - Anomalies that occur in the motion of the vehicle such as vehicle acceleration and orientation.
    \item External anomalies - Anomalies that occur in the external environment such as obstacles and external objects moving towards the vehicle.
\end{enumerate}
We also divide the anomalies into two categories based on how we can capture them; as follows:
\vspace{-0.04in}
\begin{enumerate}
    \item Instant anomalies - Anomalies which occur and can be observed in time instances, such as high or low values of acceleration and spatial anomalies in camera images.
    \item Transitional anomalies - Anomalies which occur and can be observed in transitions between subsequent time instances,  such as abnormal transitions in subsequent camera images and abnormal transitions in velocity readings.
\end{enumerate}
Fig.~\ref{fig:types1} illustrates two examples of instant anomalies which are a spatial anomaly (left) and unusual values in linear acceleration (right). The image sequence shown in Fig.~\ref{fig:types2} shows an example of a transitional anomaly, where the transition between subsequent camera frames are anomalous.

In this paper, we propose multiple self-supervised algorithms to capture all types of anomalies discussed above. In particular, to detect both internal and external anomalies we process both IMU and camera data by standalone algorithms. While IMU data are particularly useful in detecting internal anomalies, camera frame sequences can be used to detect both internal and external anomalies. To capture both instant and transitional anomalies, we propose reconstruction based and forecasting based algorithms. We assume that in the event of an instant anomaly, a model which has been trained to reconstruct the same input sample would get confused and return a high reconstruction error. Also, we assume that in the event of a transitional anomaly, a model which is trained to forecast next instance from previous instances would get confused and return a high forecasting error.

To identify anomalies in video data, the immediate-past three frames are fed into a parametric model, which captures both sequential and spatial information to predict the next frame. The next frame prediction error is compared against a threshold to determine whether the frame corresponds to an anomaly or not. We employ two approaches to detect anomalies with IMU data: the first is a Long Short Term Memory (LSTM)~\cite{hochreiter1997long} based autoencoder and the second is an LSTM based forecaster. The LSTM autoencoder reconstructs three consecutive IMU vectors and the LSTM forecaster predicts the next vector using the previous three IMU vectors. Based on the reconstruction error or the prediction error, and a threshold, timestamps are classified as anomalous or not. We implement our system using Robot Operating System (ROS)~\cite{quigley2009ros} for real-time operation. The composition of algorithms won the runner up at the IEEE Signal Processing Cup 2020 anomaly detection challenge \cite{SPcup}. 


\section{Related Work}
\label{se:related}
Video prediction has garnered considerable attention from the research community and the industry due to its application in domains such as autonomous vehicle navigation \cite{xu2017end} and robot manipulation \cite{finn2017deep}, aside from anomaly detection \cite{medel2016anomaly}, where it has been used to understand the behavior of the surrounding environment with time. In this context, sequential models such as Recurrent Neural Networks (RNNs) typically outperform non-sequential ones, as adjacent video frames often share valuable information among them, which the latter finds hard to grasp \cite{zhou2016learning,villegas2017decomposing,finn2016unsupervised,patraucean2015spatio}. The works of \cite{yuen2010data,lan2014hierarchical,kitani2012activity,huang2014action,koppula2015anticipating} demonstrate prediction of high-level information, such as the action of a person in video frames using supervised learning. The research community mainly propose models based on unsupervised learning to predict low-level details in video frames such as pixel-level information \cite{villegas2017decomposing,finn2016unsupervised}. Several works demonstrate the effective usage of different types of auto-encoders to detect anomalies in video streams. In these works, the authors reconstruct video frames with the help of autoencoders and calculate the reconstruction error between the output and the original frame to identify anomalies. Ribeiro \etal \cite{ribeiro2017} and Gutoski \etal  \cite{gutoski2017} showed that deep convolutional auto-encoder models can successfully capture high-level spatial and temporal features of video frames to detect anomalies. Furthermore, Chomg and Tay  \cite{chong2017}, and Xu \etal employed Spatio-temporal autoencoders and variational autoencoders, respectively. Further, Duman and Erdem \cite{duman2019} proposed a method using convolutional autoencoders and convolutional LSTMs to detect anomalies in which the authors use a dense optical flow to extract velocity and location information of objects. 

Several works have been presented in the context of anomaly detection of autonomous systems. Olier \etal \cite{olier2017dynamic} demonstrated how an autonomous agent can be trained to mimic human behavior using a variational deep generative architecture, and Baydoun \etal \cite{baydoun2018multi} introduced an anomaly detection model by fusing two viewpoints together; the shared layer and the private layer. Here, the shared layer observes odometry data of all the moving agents externally while the private layer is only accessible to the relevant agent and observes visual data captured by each agent. Campo \etal \cite{campo2019learning} employed a Gaussian process regression to obtain the most probable motion of an agent according to its current position and segmented the state space into spatial zones. Here, the authors employ a set of Kalman filters to track the behavior of agents in each region and to detect anomalous behavior. Iqbal \etal \cite{iqbal2019clustering} discussed the usage of an improved version of the Growing Neural Gas algorithm to cluster multi-sensory data optimally. This approach reduces computational complexity while maintaining detection accuracy. Kanapran \etal \cite{kanapram2019self} proposed a dynamic Bayesian network to evaluate and detect abnormal behavior based on internal cross-correlational parameters, using the Hellinger distance metric as the abnormality measurement. Furthermore, Ravanbakhsh \etal \cite{ravanbakhsh2018hierarchy,ravanbakhsh2018learning} proposed a novel method using an incremental hierarchy of cross-modal Generative Adversarial Networks (GANs) to process visual data obtained from a camera to build a self-awareness model and detect anomalies by computing distance between observed data and generated image frames from the GANs.

\section{Proposed Network Topologies}
\label{se:method}
The proposed architecture comprises of two separate systems to process IMU data and image data, i.e., the frames of the video, where each system identifies abnormal timestamps independently. For IMU processing, we propose two alternatives based on reconstruction and forecasting. The  reconstruction-based approach learns to auto-encode normal samples, thus for an abnormal input sample, the reconstruction error is expected to rise. The forecasting-based approach learns to predict the next sample given previous normal samples in a sequence, hence it is expected to return a high forecasting error if the inputs are abnormal. For image data processing, we propose a similar forecasting approach which is then fine-tuned using conditional adversarial training \cite{mirza2014conditional}.

\subsection{Architectures for IMU Processing}
\label{ss:IMU}

The IMU data contains linear acceleration, angular velocity, and orientation data. However, we only use linear acceleration and angular velocity for anomaly detection. Since both of them represent a rate of change in another measure (velocity, angle), high values of these sensor data are often correlated with abnormal behaviour. The IMU data vector at a given timestamp $t$ ($x_t$) is a six-dimensional vector containing the angular velocities and the linear accelerations in the three directions; 
\begin{equation}
x_t = 
    \begin{bmatrix}
    a_x & a_y & a_z & l_x & l_y & l_z\\
    \end{bmatrix}^T.
\end{equation} 
Here, $a_x$, $a_y$, $a_z$ represent angular velocities and $l_x$, $l_y$, $l_z$ represent linear accelerations in the $x$, $y$, $z$ directions, respectively. We propose two approaches to detect abnormalities which are based on reconstructing input samples and predicting the next sample from previous samples.

\begin{figure}[t]
\begin{subfigure}[b]{0.49\columnwidth}
\centering
\includegraphics[width=0.9\linewidth, height=2in]{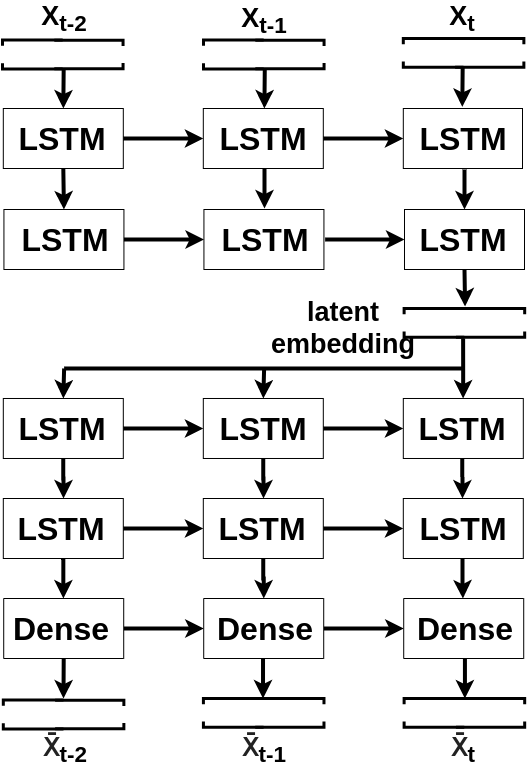}
\caption{LSTM Autoencoder}
\label{fig:lstm_auto}
\end{subfigure}
\begin{subfigure}[b]{0.49\columnwidth}
\centering
\includegraphics[width=0.9\linewidth]{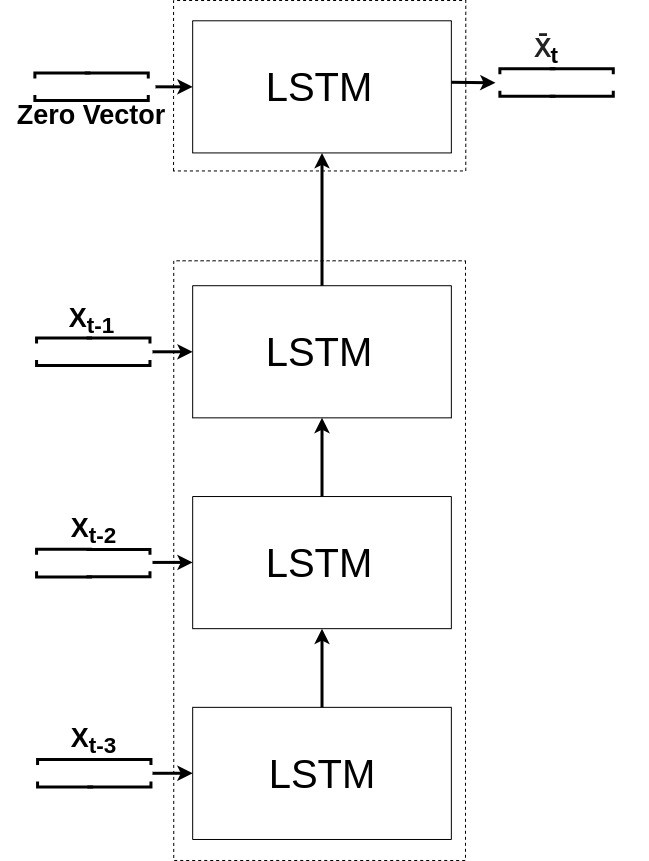}
\caption{LSTM Forecaster}
\label{fig:lstm_forecaster}
\end{subfigure}
\vspace{-0.05in}
\caption{(a) LSTM Autoencoder where input samples in a sliding window are reconstructed  (b) LSTM Forecaster which predicts the fourth IMU vector from the three consecutive IMU vectors.}
\label{fig:IMU_nets}
\vspace{-0.05in}
\end{figure}

\subsubsection{LSTM Autoencoder}
\label{sss:lstmauto}

We propose the use of autoencoders to learn to reconstruct the normal data. The Autoencoder learns a basis representation of the normal data and to reconstruct them with minimal error, allowing the reconstruction error to be used as an anomaly metric. To make the reconstruction process smooth over the past samples, we incorporate an LSTM~\cite{hochreiter1997long} architecture to expand the autoencoder to aid from previous samples as well. We base ourselves on the assumption that the reconstruction error is high for data points that lie in significantly different ranges from the data points the model has seen. 

Fig.~\ref{fig:lstm_auto} illustrates the LSTM Autoencoder architecture which consists of two parts: the encoder and the decoder. The encoder receives three IMU data vectors of three consecutive timestamps. The first LSTM layer outputs 128-dimensional feature vectors and the second LSTM layer reduces the feature size to 64, where the final time step of the second layer outputs a 64-dimensional encoded embedding. This encoded vector is then repeated three times and fed to the three cells of the first LSTM layer of the decoder which outputs three 64-dimensional feature vectors. The next layer increases the feature size to 128, finally a time distributed dense layer provides output with the same dimensions as the input. The reconstructed outputs are then compared with the targets which are the inputs themselves to calculate the mean squared error to back-propagate. This process is repeated over such windows of three samples which slide through the normal sequences.

\subsubsection{LSTM Forecaster}
\label{sss:forecasting}

Here, we propose the use of encoder-decoder recurrent models which are conventionally used for sequence-to-sequence prediction problems \cite{sutskever2014sequence}. Given a sequence of IMU vectors of consecutive timestamps, the recurrent encoder processes the sequential data to return a latent hidden state which is then fed to the recurrent decoder to predict the IMU vectors of the next samples. We base ourselves on the assumption that once the network has learned to predict future samples from given normal samples, the next sample prediction gets confused and deviates from the actual for any abnormality in the transitions in the input sequence, hence leads to high prediction error. 

Fig.~\ref{fig:lstm_forecaster} shows the architecture of the LSTM~\cite{hochreiter1997long} based Forecaster. Given three IMU vectors from three consecutive timestamps to the encoder which is an LSTM layer with three steps, it processes the sequential data and returns the latent hidden state of the final timestep. The decoder, which is a single LSTM cell, is then initiated with this hidden state and a zero vector as the input, to predict the IMU vector of the next sample. Although we can extend the decoder to predict several future samples, we limit to one sample and exclude such investigation in this paper. The predicted sample is then compared against the actual future sample to calculate the mean squared error which is then back-propagated. During inference, this error is expected to rise for any abnormality in the transitions in the input sequence.

\subsection{Architecture for Image Processing}
\label{ss:method_images}

To capture anomalous behaviour from images, it is important to process them sequentially to capture sudden abnormal transitions between consecutive frames. Once the network successfully learns to predict the future frame from the previous frames, during the inference time if there are sudden unpredictable transitions between consecutive frames it becomes hard to predict future frames, hence the system gives higher prediction errors. Here, we use a convolutional neural network based sequential model (CNN-LSTM Forecaster) to predict the next frame from the immediate-past frames. This is similar to the LSTM Forecaster introduced for IMU processing (Sec. \ref{sss:forecasting}). However, the input frames are pre-processed by a convolutional encoder to reduce the dimensions, and the LSTM prediction is post-processed by a convolutional decoder to construct the future frame. The predicted frame and the actual occurrence are then compared to determine the prediction error which is minimized. 

\begin{figure}[t]
\begin{subfigure}[c]{0.49\columnwidth}
\includegraphics[width=\linewidth]{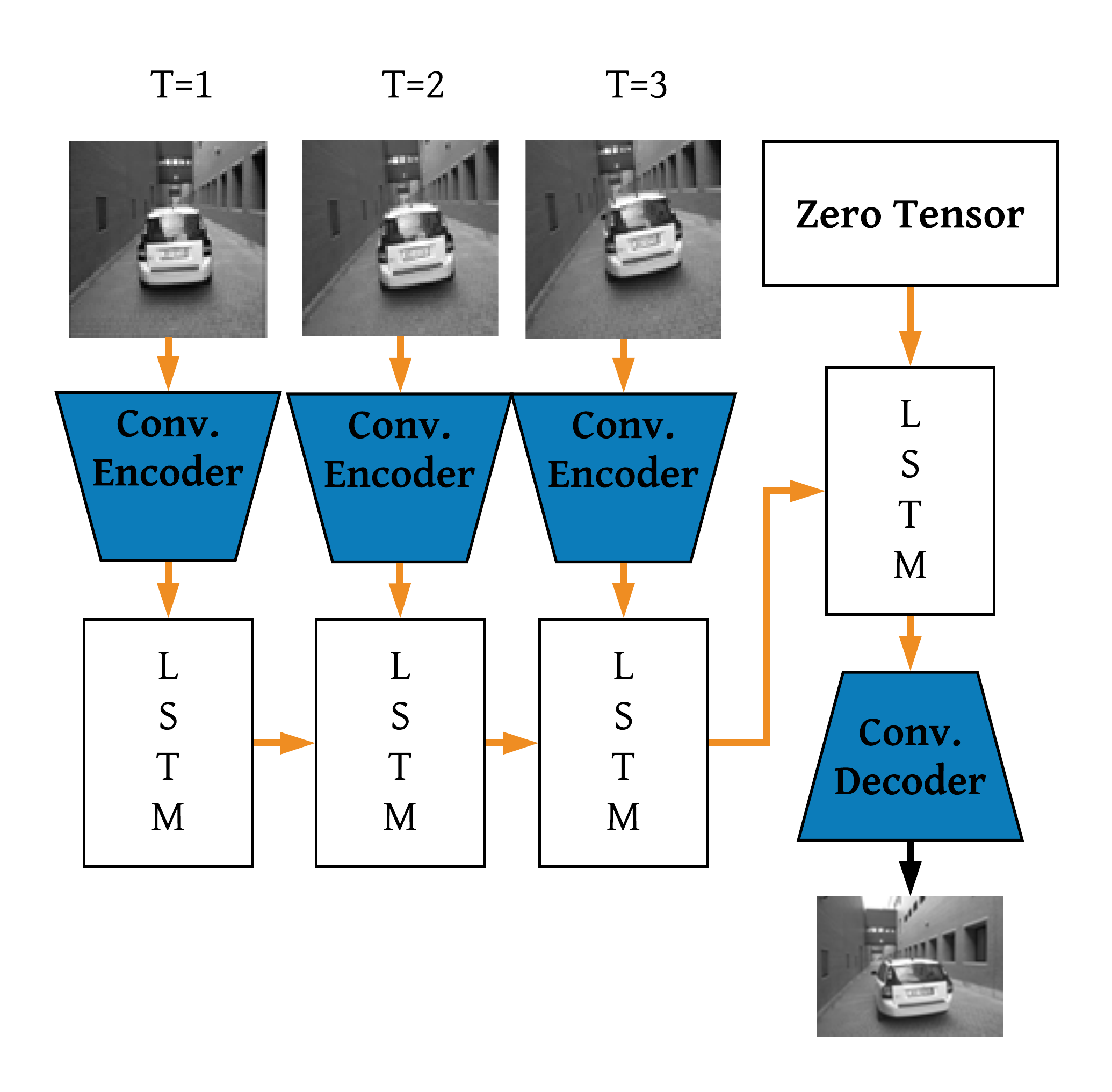}
\caption{CNN-LSTM Forecaster}
\label{fig:cnn-lstm-forecaster}
\end{subfigure}
\begin{subfigure}[c]{0.49\columnwidth}
\includegraphics[width=\linewidth, height=1.7in]{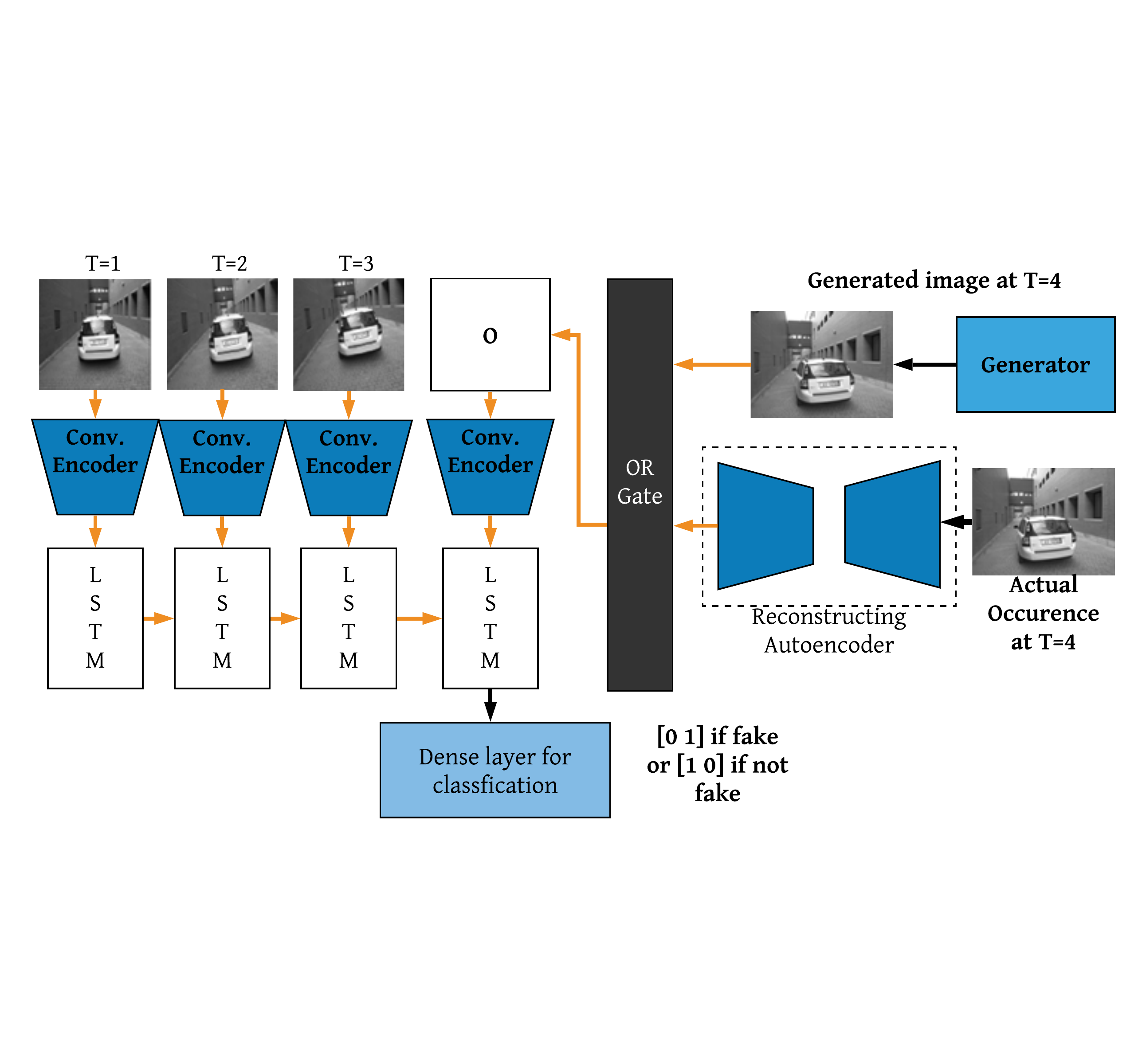}
\caption{CGAN Discriminator}
\label{fig:disc}
\end{subfigure}
 \vspace{-0.05in}
\caption{(a) A sequence of previous frames is fed into the LSTM Forecaster via convolutional encoders. The forecaster's latent prediction is fed to the convolutional decoder to construct the predicted frame.
(b) A sequence of four frames in which the fourth one is either generator output or the reconstructed version of the real occurrence, are fed to the LSTM via convolutional encoders. The LSTM then classifies the fourth frame as fake (generated) or real.}
\label{fig:Image_nets}
\vspace{-0.05in}
\end{figure}

Fig.~\ref{fig:cnn-lstm-forecaster} shows the CNN-LSTM Forecaster. For the convolutional encoder and decoder networks, we use the SegNet architecture. The encoder, which consists of nine hidden convolutional layers takes a $128\times 128 \times 1$ image and produces a $4\times 4\times 64$ latent tensor. This is flattened and fed to the LSTM Forecaster which takes three such embeddings and predicts the fourth frame's latent representation. Once the LSTM Forecaster produces the future frame embedding, it is reshaped to $4\times 4\times 64$ and the convolutional decoder reverses the process of the encoder to construct a $128\times 128 \times 1$ image at the output. The decoder consists of convolutional layers which are followed by up-sampling layers where the dimensions are increased. All hidden layers in both encoder and decoder use leaky ReLU activation with 0.2 slope, except the final layers which use tanh activation. 

To make the CNN-LSTM Forecaster more robust to unseen data and to prevent the models overfitting to the relatively small dataset, we further deploy a CGAN approach~\cite{mirza2014conditional}. Here, the trained Forecaster is used as the generator where another CNN-LSTM architecture is used as the Discriminator. The Discriminator, as shown in Fig.~\ref{fig:disc}, uses a four-step LSTM where the first three steps take the three input frames' latent representations via a convolutional encoder. The fourth step to the LSTM is either the encoded Forecaster generated image or the real fourth image. The discriminator is supposed to distinguish between the real and generated fourth frame. We feed the real fourth frame through convolutional encoder and decoder to subject them to similar information loss as generated images otherwise, the discriminator would easily overpower the generator.   

\section{Data Preparation, Training and Inference}
\subsection{Training with IMU Data}
\label{ss:exp_imu}

We use only the linear acceleration and angular velocity values of the IMU data as explained in Section \ref{ss:IMU}. The dataset contains six normal scenarios and six abnormal scenarios that are given in ROS bag files. We refer to the normal scenarios as normal-0, normal-1, normal-2, normal-3, normal-4, and normal-5. The six abnormal bag files are referred to similarly. The LSTM Autoencoder and LSTM Forecaster are trained only on four of the normal bag files (normal-1,...,4) where we use normal-0 for finding the threshold, and normal-5 along with all abnormal cases for testing.  

The training dataset consists of 551 IMU vectors with each containing six features. As a preprocessing step, each feature is scaled to the range [-1,1]. We create two distinct datasets for the two models. For the LSTM Autoencoder, we use a sliding window of three consecutive vectors. The training dataset contains 549 such sets. In this case the input and targets to the model are the same. To train the LSTM Forecaster, we use the same sliding window of length three with the target being the fourth frame which follows the three frames in every window. There are 548 such sets in this training dataset. The normal-0 which is used for thresholding contains 302 timestamps which leads to 298 4-frame segments for the LSTM Forecaster and 299 3-frame segments for the LSTM Autoencoder. Each model is trained in the respective training dataset for 500 epochs with a learning rate of 0.01. We use mean squared error as the loss function and a batch size of 1.

\subsection{Training with Image Data}
\label{ss:exp_image}

The image dataset consists of sequences of RGB images taken from the drone’s frontal camera at fixed time intervals. The images are from six normal scenarios and six abnormal scenarios as mentioned in Section \ref{ss:exp_imu}. These images are converted to gray-scale and resized to $128 \times 128$. Furthermore, the image pixel values are normalized to the range of [-1, 1]. We augment the normal scenarios with horizontal flipping which gives us six more normal scenarios which are the mirrors of the original six normal scenarios. 

From each normal scenario, we construct sequences of four images where the first three images are inputs and the fourth image is the target for prediction. We construct 810 such sequences by sliding the length-4 window through all normal scenarios.  We annex all the image sequences together and randomly pick 100 image sequences as the threshold determination set and 100 more are allocated for testing. The rest of the 610 image sequences are used to train the model. The 100 test sequences are later combined with 196 sequences obtained similarly from the abnormal scenarios to build the total test set of 296 four-image segments.

First, we combine the convolutional encoder and decoder parts together which resembles the SegNet \cite{segnet} architecture and train for the image reconstruction task. We use the individual images in the training set which are augmented using horizontal flips, random rotation within ten degrees, width and height shifts and zoom. The trained encoder and decoder parts are then plugged into the CNN-LSTM Forecaster (Fig.~\ref{fig:lstm_forecaster}) which is then trained to predict the fourth frame from 3 consecutive input frames. During this phase, the weights of the convolutional encoder and decoder are frozen, and only the LSTM cells are learned for the prediction task. For both encoder-decoder training and CNN-LSTM Forecaster training, we use the addition of the mean squared error and the mean absolute error as the loss function. Such an addition enables the model to benefit from stable convergence while being robust to outliers. In both cases, we train the models for 100 epochs where the learning rate is initialized as 0.001 and decayed by a factor of 10 after 50 and 80 epochs respectively. 

Afterward, the trained LSTM Forecaster is used as the generator in a CGAN \cite{mirza2014conditional} where the LSTM Forecaster's next frame prediction is fine-tuned with the adversarial loss along with the prediction loss. Section \ref{ss:method_images} and Fig.~\ref{fig:disc} explains the Discriminator architecture we use for this purpose. During this phase, we let all the weights of the Forecaster to be freely updated, and we use a combined loss function of both prediction loss and the adversarial loss as proposed in \cite{pix2pix}.

\subsection{Flagging Anomalies During Inference}
\label{ss:exp_inference}

\begin{figure}[t]
\centering
\includegraphics[width=0.9\linewidth]{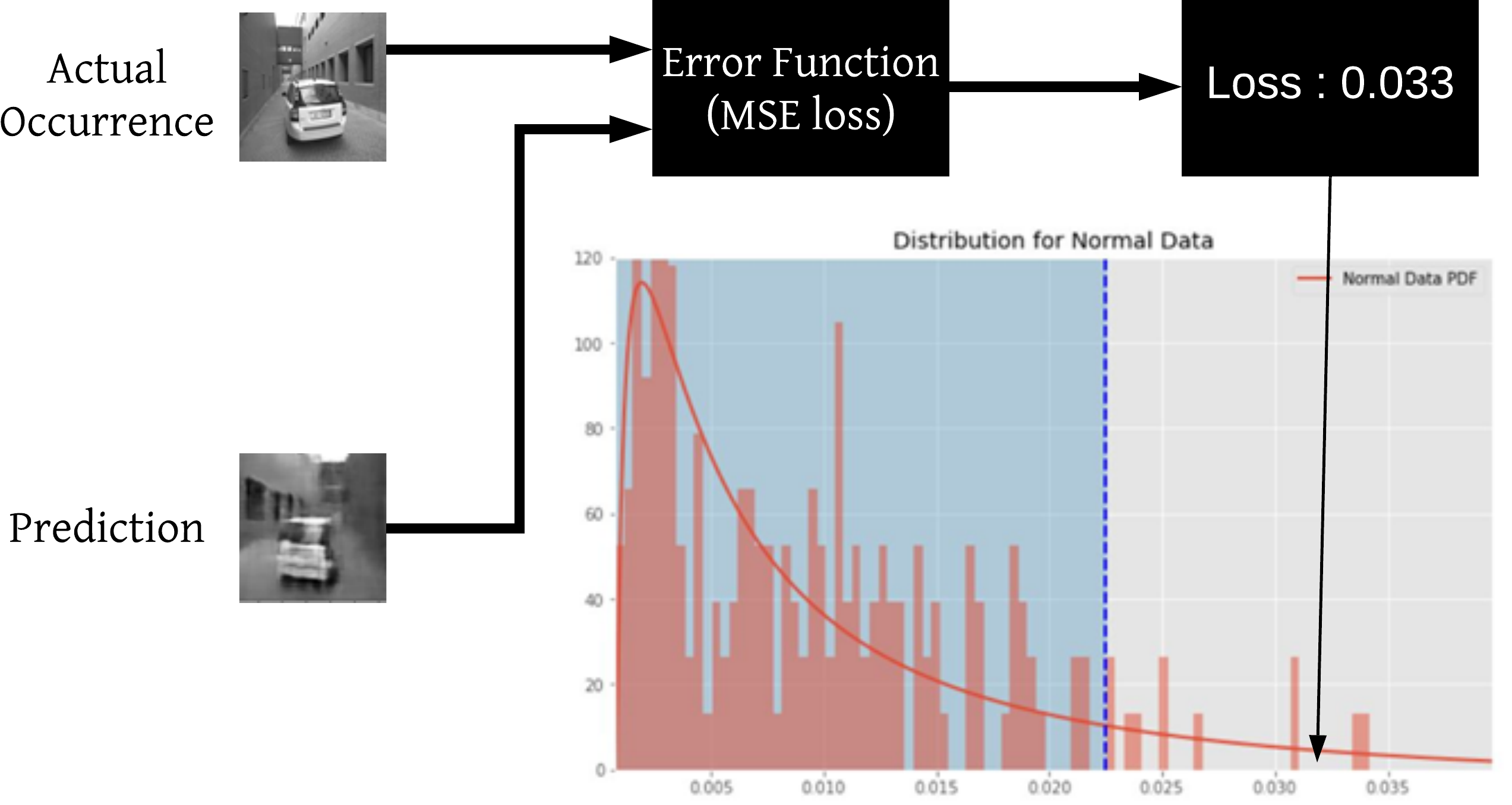}
\vspace{-0.05in}
\caption{We record the prediction error in the thresholding set and compute the histogram and the probability distribution. We fit a statistical curve to the distribution and select the threshold at 95\% right-tailed confidence. During inference, if the error exceeds the threshold we predict the timestep as abnormal.} 
\label{fig:normcatch}
\vspace{-0.05in}
\end{figure}

\begin{figure}[ht]
 \centering
 \begin{subfigure}[b]{0.49\columnwidth}
 \includegraphics[width=0.95\linewidth, height=1in]{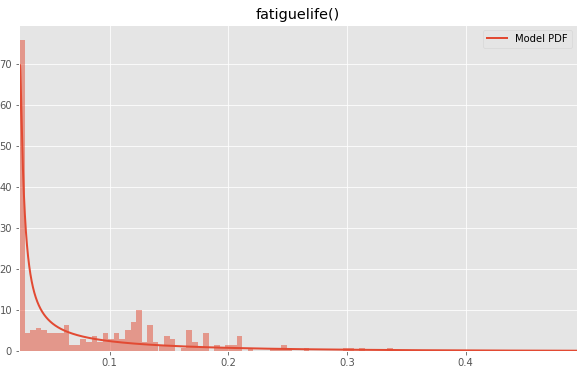}
 \caption{$e_a$ Autoencoder}
 \label{fig:ea_autoencoder}
 \end{subfigure}
 \begin{subfigure}[b]{0.49\columnwidth}
 \includegraphics[width=0.95\linewidth, height=1in]{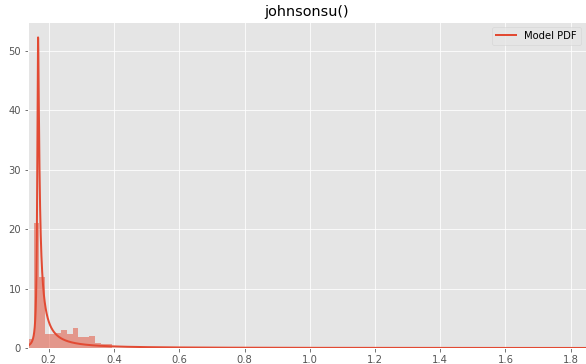}
 \caption{$e_l$ Autoencoder}
 \label{fig:el_autoencoder}
 \end{subfigure}
 \begin{subfigure}[b]{0.49\columnwidth}
 \includegraphics[width=0.95\linewidth, height=1in]{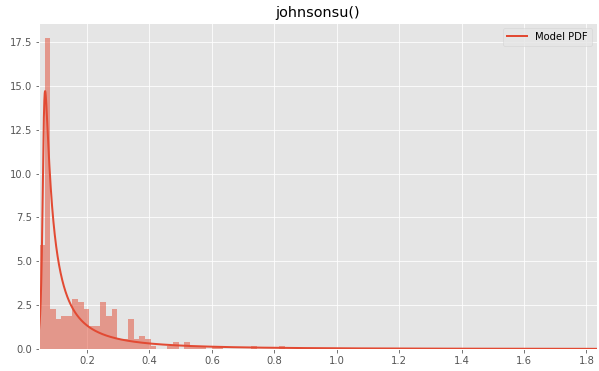}
 \caption{$e_a$ Forecaster}
 \label{fig:ea_forecaster}
 \end{subfigure}
 \begin{subfigure}[b]{0.49\columnwidth}
 \includegraphics[width=0.95\linewidth, height=1in]{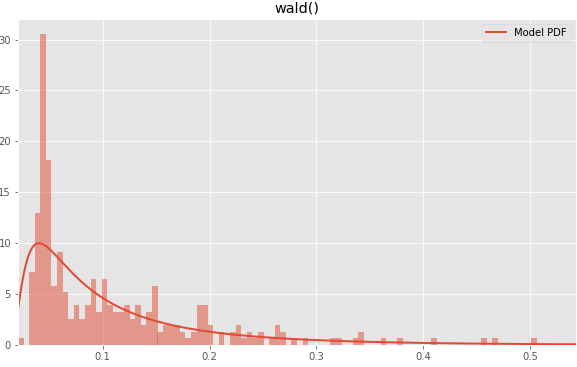}
 \caption{$e_l$ Forecaster}
 \label{fig:el_forecaster}
 \end{subfigure}
 \vspace{-0.05in}
 \caption{By having separate thresholds, at a given timestamp, we can identify whether the vehicle has an anomaly in angular velocity or linear acceleration.}
 \label{fig:imu_curves}
 \vspace{-0.05in}
\end{figure} 
\label{se:experiments}

Both models used for IMU data processing as well as the model used for image data processing are based on either the reconstruction of the same sample or the prediction of future samples. Therefore, we record the reconstruction or prediction error of each model in the thresholding dataset and compute error histograms that are converted to probability distributions. We fit statistical distributions to the derived probability distributions. To determine the best-fit curve for each case, we use the Kolmogorov-Smirnov test \cite{kolmogorov-smirnov}. Once a statistical curve is fit for each distribution, we set the thresholds at the 95\% right-tailed confidence which is then used to flag abnormal behaviour in the test set. In particular, during inference, if the reconstruction/prediction error is above this threshold we flag the timestep as abnormal (See Fig.~\ref{fig:normcatch}).

In IMU data processing, the reconstruction/prediction error for a particular step is a six-dimensional vector containing angular velocities and linear acceleration each in three directions ([$e_{a_x} e_{a_y} e_{a_z} e_{l_x} e_{l_y} e_{l_z}$]). Here, we plot histograms and compute thresholds for the angular velocities and linear acceleration separately. In particular, we define the error related to angular velocities $e_a$ as the mean of the 3 angular velocity errors and the error related to linear acceleration $e_l$ as the mean of the 3 linear acceleration errors. The reason for such division is that angular velocity and linear acceleration are two distinct measures which might not be correlated always.

For the LSTM Autoencoder, $e_a$ is fit with a Birnbaum-Saunders distribution \cite{fatiguelife} (parameters - c: 2.053, location: 0.022, scale: 0.019) and $e_l$ is fit with a Johnson's SU distribution \cite{johnsu} (parameters - a: 0.89,b: 0.44,location: 0.16,scale: 0.0024). The 95\% right-tailed confidence thresholds for the $e_a$ and $e_l$ are computed as 0.276 and 0.531 respectively. Similarly the thresholds of $e_a$ and $e_l$ in the LSTM Forecaster are calculated as 0.655 and 0.322. Fig.~\ref{fig:imu_curves} shows these distributions where $e_a$ and $e_l$ distributions in each case show clear differences from each other. For the vision system, the prediction error is fit with the Normal Inverse Gaussian distribution \cite{norminvgauss} (parameters - a : 0.326, b: 0.291, location :0.061, scale: 0.01). The 95\% right-tailed confidence threshold is computed as 0.1598.

\section{Results and Discussion}
\label{se:results}

\begin{table}[t]
\centering
\caption{Comparison of the two models for the IMU data on the abnormal bags with hand-labeled ground truth.}
\vspace{-0.05in}
\label{tab:imu}
\begin{tabular}{@{}c|c|c|c|c|c|c@{}}
\hline
Bagfile         & \multicolumn{2}{c|}{Precision}  &  \multicolumn{2}{c|}{Recall}   & \multicolumn{2}{c}{F1-score}   \\ \cline{2-7}
                    & Auto.          & Fore.         & Auto.         & Fore.         & Auto.         & Fore.  \\ \hline
abnormal-0          & 0.92           & 0.92          & 1.00          & 1.00          & 0.96          & 0.96   \\ 
abnormal-1          & 0.99           & 0.98          & 1.00          & 0.98          & 1.00          & 0.98   \\ 
abnormal-2          & 0.96           & 0.96          & 0.99          & 0.99          & 0.97          & 0.97   \\ 
abnormal-3          & 0.98           & 0.99          & 1.00          & 0.99          & 0.99          & 0.99   \\ 
abnormal-4          & 1.00           & 0.99          & 0.99          & 1.00          & 1.00          & 1.00   \\ 
abnormal-5          & 1.00           & 1.00          & 0.96          & 0.98          & 0.98          & 1.00   \\ \hline
average             & 0.97           & 0.97          & 0.99          & 0.99          & 0.98          & 0.98   \\ \hline
\end{tabular}
\vskip-2ex
\end{table}

\begin{table}[ht]
\caption{Performance of CNN-LSTM Forecaster with only prediction error and after fine-tuning with adversarial loss which further improves the recall, accuracy and F1 score.}
\vspace{-0.05in}
\begin{center}
\begin{tabular}{c|c|c|c|c}
\hline
Metrics & Precision & Recall & Accuracy & F1\\
\hline
Only Prediction Loss & 0.9381 & 0.9081 & 93.81\% & 0.9518  \\ 
Prediction Loss + CGAN & 0.9269 & 0.9821 & 94.18\% & 0.9537 \\ 
\hline
\end{tabular}
\end{center}
\label{tab:cgan}
\vskip-2ex
\end{table}

\begin{figure}[ht]
\begin{subfigure}[b]{0.19\columnwidth}
\includegraphics[width=\linewidth]{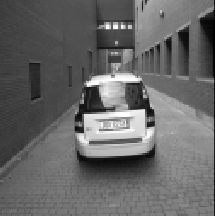}
\caption{\scriptsize{1}}
\end{subfigure}
\begin{subfigure}[b]{0.19\columnwidth}
\includegraphics[width=\linewidth]{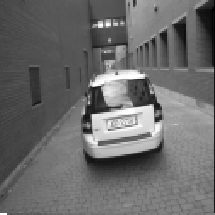}
\caption{\scriptsize{2}}
\end{subfigure}
\begin{subfigure}[b]{0.19\columnwidth}
\includegraphics[width=\linewidth]{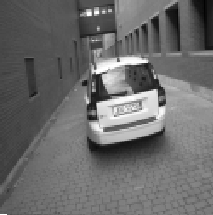}
\caption{\scriptsize{3}}
\end{subfigure}
\begin{subfigure}[b]{0.19\columnwidth}
\includegraphics[width=\linewidth]{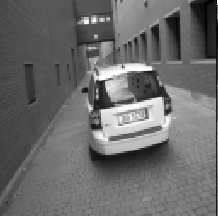}
\caption{\scriptsize{4-Real}}
\end{subfigure}
\begin{subfigure}[b]{0.19\columnwidth}
\includegraphics[width=\linewidth]{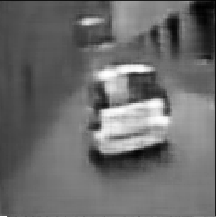}
\caption{\scriptsize{4-Predicted}}
\end{subfigure}
 \vspace{-0.05in}
\caption{Performance of the CNN-LSTM Forecaster on a normal sequence. The fourth frame is successfully predicted by the pattern shown in the previous frames.}
\label{fig:normal_case}
\vspace{-0.05in}
\end{figure}

\begin{figure}[ht]
\begin{subfigure}[b]{0.19\columnwidth}
\includegraphics[width=\linewidth]{an_a_1.png}
\caption{\scriptsize{1}}
\end{subfigure}
\begin{subfigure}[b]{0.19\columnwidth}
\includegraphics[width=\linewidth]{an_a_2.png}
\caption{\scriptsize{2}}
\end{subfigure}
\begin{subfigure}[b]{0.19\columnwidth}
\includegraphics[width=\linewidth]{an_a_3.png}
\caption{\scriptsize{3}}
\end{subfigure}
\begin{subfigure}[b]{0.19\columnwidth}
\includegraphics[width=\linewidth]{an_a_actual.png}
\caption{\scriptsize{4-Real}}
\end{subfigure}
\begin{subfigure}[b]{0.19\columnwidth}
\includegraphics[width=\linewidth]{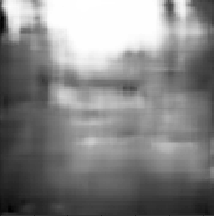}
\caption{\scriptsize{4-Predicted}}
\end{subfigure}
 \vspace{-0.05in}
\caption{Anomaly Type 01: Due to the unpredictable movement of the drone, there are sudden unpredictable transitions in the sequence. Thus the prediction is fully corrupted.}
\label{fig:anomal_case1}
\vspace{-0.05in}
\end{figure}

\begin{figure}[ht]
\begin{subfigure}[b]{0.19\columnwidth}
\includegraphics[width=\linewidth]{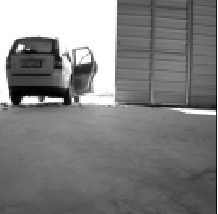}
\caption{\scriptsize{1}}
\end{subfigure}
\begin{subfigure}[b]{0.19\columnwidth}
\includegraphics[width=\linewidth]{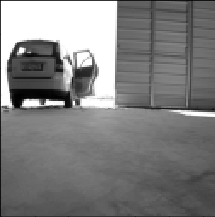}
\caption{\scriptsize{2}}
\end{subfigure}
\begin{subfigure}[b]{0.19\columnwidth}
\includegraphics[width=\linewidth]{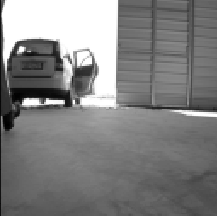}
\caption{\scriptsize{3}}
\end{subfigure}
\begin{subfigure}[b]{0.19\columnwidth}
\includegraphics[width=\linewidth]{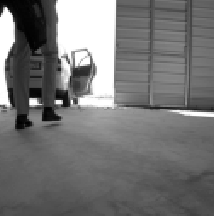}
\caption{\scriptsize{4-Real}}
\end{subfigure}
\begin{subfigure}[b]{0.19\columnwidth}
\includegraphics[width=\linewidth]{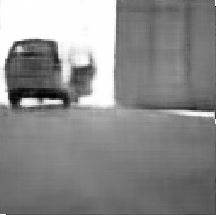}
\caption{\scriptsize{4-Predicted}}
\end{subfigure}
\vspace{-0.05in}
\caption{Anomaly Type 02: Although the observing agent is stable and the first 3 frames look normal, the sudden appearance of a man in the fourth frame is not predicted by the model; hence, the high prediction error.}
\label{fig:anomal_case2}
\vspace{-0.05in}
\end{figure}

To evaluate the IMU models we use the test set which contains one normal case (normal-5) and all the abnormal cases. We establish ground-truth by classifying each data point in the test set as an anomaly or not. For linear accelerations along each orthogonal axis, maximum and minimum values corresponding to normal data are used as thresholds to detect anomalies. If acceleration along any of the axes is found to be beyond the threshold, we label them as abnormal. Angular velocities are labeled in a similar manner. We assume such irregularities in angular velocity and linear acceleration are correlated with abnormal movement. 

The normal-5 scenario is not used for the training or thresholding processes, hence we first evaluate our models in this scenario where the  LSTM Autoencoder achieves an accuracy of 95.3\% and the LSTM Forecaster achieves an accuracy of 100\% in predicting whether a particular step is normal. Table \ref{tab:imu} illustrates the performance of the two models in abnormal scenarios where both perform equally well.

To evaluate the image sequence anomaly detection process, we report the performance of the CNN-LSTM Forecaster on a test set. The test set contains 296 frame sequences taken from six abnormal scenarios and unused sequences from the normal cases. We establish test set ground-truth by manually observing each sequence of the four consecutive frames in a sliding window, determining whether the fourth frame is abnormal. Here, we specifically pay attention to the ability to predict the fourth frame given the three previous frames. If the initial three frames in a particular segment show unpredictable transitions, or fourth frame deviates significantly from the pattern followed by its preceding three frames, we label the fourth frame as an anomaly.

Table \ref{tab:cgan} compares the CNN-LSTM Forecaster's performances when trained with only the prediction loss and when further fine-tuned using conditional adversarial loss. When the Forecaster is fine-tuned with the adversarial loss, the recall, accuracy and the F1 score are improved. We further plot handpicked frame sequences representing both normal and abnormal segments from the test set. Fig.~\ref{fig:normal_case} shows a normal frame segment where the fourth frame is successfully predicted given the three previous frames. Fig.~\ref{fig:anomal_case1} shows a segment captured by a drone which is already in an anomalous movement. In such a case, it is difficult for the system to predict the future frame. In the frame segment shown in Fig.~\ref{fig:anomal_case2}, the model fails to predict the sudden appearance of the man in the fourth frame. During such sudden unpredicted occlusions, the IMU sensors continue to give normal reading since the movement of the vehicle is regular. Hence, these types of anomalies are not flagged by the IMU processing.

\section{Conclusion}
\label{se:conslusion}
In this paper, we proposed self-supervised deep learning algorithms to detect anomalies in autonomous systems using video and IMU data. Our algorithms were based on reconstructing the same input sample or predicting the next sample from a given sequence. We expected that for input samples that are significantly dissimilar to the samples that the reconstruction model has seen during training, the reconstruction error tends to rise. Additionally, if there are unpredicted, sudden transitions in a given sequence of samples, it is hard to predict the next sample. Consequently, the prediction error rises. The proposed models for IMU-based anomaly detection achieve an accuracy of 91\% and an F1-score of 0.99 whereas the proposed model for image-based anomaly detection achieves an accuracy of 94\% and an F1-score of 0.95.

\bibliographystyle{IEEEtran}
\balance
\bibliography{IEEEabrv, IEEEexample}

\end{document}